\documentclass[runningheads]{llncs}
\usepackage[T1]{fontenc}
\usepackage{graphicx,verbatim}
\usepackage{amsmath}
\usepackage{amssymb}
\usepackage{multirow}
\usepackage{threeparttable}
\usepackage{xcolor}
\usepackage{microtype}
\newcommand{\myparagraph}[1]{\vspace{1pt}\noindent{\bf{#1}}}
\definecolor{myblue}{RGB}{0,0,200}  
\usepackage[colorlinks,linkcolor=red, anchorcolor=myblue, citecolor=myblue, urlcolor=magenta]{hyperref}

\begin{document}
\title{NeuroSonic: Conditional Flow Matching for EEG-to-Speech Reconstruction}
\titlerunning{NeuroSonic: EEG-to-Speech Reconstruction}
%

\author{%
Wenhao Gao\inst{1}\thanks{Equal contribution.}\and
Yifan Wang\inst{1}\protect\footnotemark[1]\and
Yijia Ma\inst{2} \and 
Carl Yang\inst{3} \and
Wen Li\inst{2} \and \\
Chenyu You\inst{1}\thanks{Corresponding author.}%
}
\titlerunning{NeuroSonic: Conditional Flow Matching for EEG-to-Speech Reconstruction}
\authorrunning{W. Gao et al.}
\institute{%
Stony Brook University, Stony Brook, NY, USA   \\ 
\email{chenyu.you@stonybrook.edu} \and
University of Texas Health Center at Houston, Houston, TX, USA \and
Emory University, Atlanta, GA, USA
}

\maketitle
\begin{abstract}
Reconstructing continuous speech from scalp electroencephalography (EEG) remains fundamentally challenging. EEG provides a weak, spatially diffuse, and highly variable measurement of distributed cortical activity, whereas speech is organized as a coherent acoustic trajectory with strong harmonic and temporal structure. The resulting mismatch makes waveform regression unstable and causes stochastic multi-step generation to be sensitive to artifact-dependent conditioning and subject variability.
We introduce \textbf{NeuroSonic}, a conditional flow-matching framework for EEG-to-speech reconstruction. Instead of predicting waveforms directly or refining them through stochastic denoising, NeuroSonic learns a deterministic probability-flow velocity field that transports a noise-corrupted acoustic state toward clean speech under EEG conditioning. EEG and audio are embedded into a shared token space and processed by a time-conditioned gated Transformer that parameterizes the transport ordinary differential equation. This formulation models trajectory evolution explicitly while avoiding iterative stochastic sampling.
We evaluate NeuroSonic on the CineBrain and EAV benchmarks under cross-subject evaluation. Across both datasets, the proposed method improves distributional realism, spectral fidelity, and perceptual quality over representative GAN-, diffusion-, and mean-flow baselines, with up to a 26.3\% gain in overall perceptual quality.
The performance gap is most evident in artifact-heavy segments, where conditioning variability is strongest. These findings indicate that deterministic conditional transport provides a stable and effective formulation for EEG-driven speech reconstruction. Code is available at \href{https://github.com/Y-Research-SBU/NeuroSonic/}{here}.


\end{abstract}

\section{Introduction}

Reconstructing continuous speech from scalp electroencephalography (EEG) entails coupling two signals with markedly different structure. EEG recordings are low-amplitude, spatially diffuse projections of distributed cortical sources~\cite{nunez2006electric,brechet2019capturing}. They exhibit substantial variability across subjects and sessions and are susceptible to motion and physiological artifacts~\cite{xu2020cross,ma2022large,vo2026inter}. In contrast, speech evolves along a highly organized acoustic trajectory characterized by harmonic structure and temporal coherence. The mapping from neural measurements to acoustic realizations is therefore indirect, temporally misaligned, and strongly confounded by nuisance variability. Although EEG-based systems have achieved promising results for constrained vocabulary classification~\cite{lee2023towards}, reconstructing natural, continuous speech with high fidelity remains unresolved. Recent EEG foundation models improve transferable representations \cite{chen2024eegformer}, but continuous speech reconstruction remains unresolved.

Generative modeling offers a principled alternative to discrete decoding~\cite{liu2021aligning,chen2021self,chen2021adaptive,you2024calibrating,you2025uncovering}. However, prevailing paradigms do not fully align with  scalp EEG. GAN-based synthesis can become unstable when the conditioning signal is weak or highly variable~\cite{goodfellow2014generative,han2023medgen3d}. Diffusion models improve optimization behavior but rely on multi-step stochastic sampling and assume a consistent corruption schedule across timesteps~\cite{ho2020denoising,ma2023pre,sun2025ouroboros,ren2025scale}. Under EEG conditioning, these assumptions are challenged by artifact-dependent noise patterns and inter-subject heterogeneity, which can accumulate across sampling steps and degrade reconstruction consistency.

\begin{figure}[t]
\centering
\includegraphics[width=0.98\linewidth]{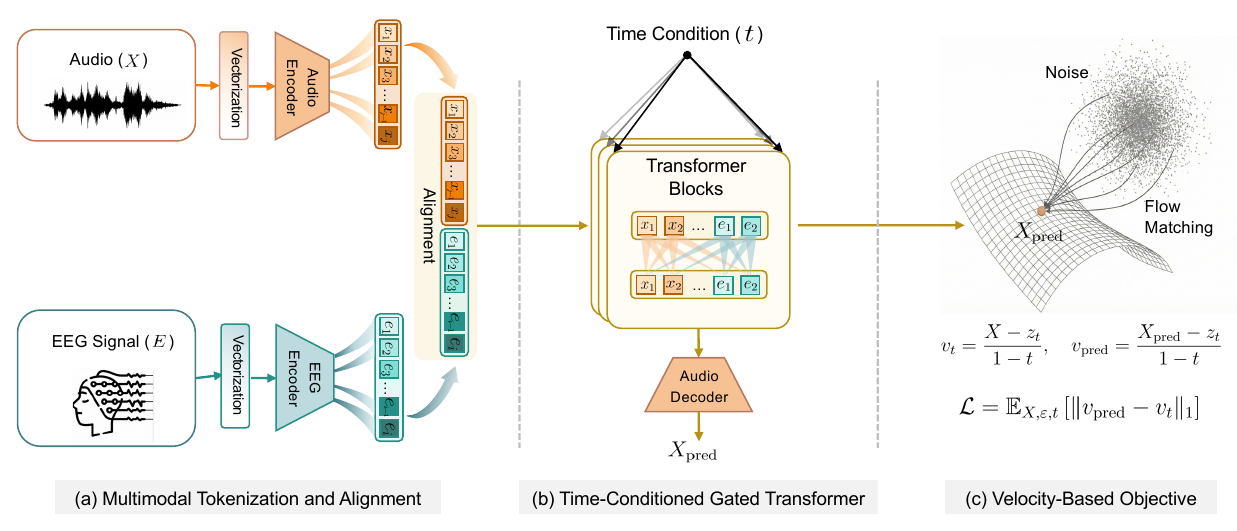}
\vspace{-5pt}
\caption{\textbf{Overview of NeuroSonic.} 
(a) EEG and audio signals are partitioned into patches, $\{E_i\}$ and $\{X_j\}$, and projected through modality-specific encoders $f_E(\cdot)$ and $f_A(\cdot)$ into a shared latent space for joint modeling. 
(b) A time-conditioned gated Transformer processes the combined sequence together with a corrupted acoustic state $z_t$, obtained by interpolating clean audio with Gaussian noise $\epsilon$ at time $t$, along the flow-matching path. Adaptive layer normalization and RMS-stabilized attention are used to preserve stable feature scaling across interpolation times.
(c) The velocity-based objective trains the predicted velocity $v_{\mathrm{pred}}$, computed from the predicted clean state $X_{\mathrm{pred}}$, to match the target transport velocity $v_t$ governing acoustic transport under EEG conditioning.
}
\label{fig:overview}
\vspace{-10pt}
\end{figure}

These motivate us to seek a formulation that models acoustic trajectory evolution directly while remaining stable under heterogeneous conditioning. Flow Matching (FM) provides a continuous-time generative framework in which a neural network learns a velocity field transporting a probability path between distributions~\cite{lipman2022flow}. Recent work explores rectified-flow-based latent synthesis to align EEG and speech representations for speech-driven clinical analysis~\cite{xiang2026cross}.
 By parameterizing deterministic probability flows rather than stochastic refinement chains, FM removes the need for iterative denoising and enables conditioning to act on the transport dynamics themselves. Recent EEG generation work further suggests that flow matching is effective for preserving continuous temporal and spectral structure in neural signals \cite{wang2026let}. This perspective is suited to speech reconstruction, where temporal coherence is intrinsic to the signal structure.

In this work, we formulate EEG-to-speech reconstruction as conditional acoustic transport. Instead of predicting waveforms in a single step or refining them through stochastic sampling, we learn a deterministic velocity field that maps corrupted acoustic states toward clean speech under EEG conditioning. Building on this formulation, we introduce \textbf{NeuroSonic}. As illustrated in Fig.~\ref{fig:overview}, EEG and audio signals are partitioned into patch-level representations and embedded into a shared latent space. A time-conditioned gated Transformer processes the joint sequence to parameterize the probability-flow ordinary differential equation governing acoustic evolution. This design enables global cross-modal interaction while stabilizing feature dynamics across interpolation times, leading to robust reconstruction under artifact corruption and cross-subject variability.
(1) We reformulate EEG-to-speech reconstruction as a deterministic, trajectory-aware inverse problem via conditional flow matching~\cite{lipman2022flow}.
(2) We propose a multimodal tokenization scheme and a time-conditioned Transformer architecture that align neural representations with acoustic dynamics within a shared latent space. 
(3) We demonstrate consistent improvements over representative GAN-, diffusion-, and mean-flow baselines on public EEG-audio benchmarks, particularly under cross-subject evaluation and artifact-heavy conditions.

\section{Method} 

\subsection{Preliminary: Flow Matching for Conditional Transport}
Flow Matching formulates generative modeling as learning a continuous-time transport between probability distributions~\cite{lipman2022flow}. Let $p_0$ denote a simple prior and $p_1 = p_{\text{data}}$ the target distribution. FM defines a probability path $\{p_t(x)\}_{t \in [0,1]}$ connecting the two and learns a velocity field that transports samples along this path. Under the linear interpolation path,
\begin{equation}
x_t = (1-t)x_0 + t x_1,
\qquad
v(x_0,x_1,t) = x_1 - x_0,
\qquad
\frac{\mathrm{d} x_t}{\mathrm{d} t} = v_\theta(x_t,t),
\end{equation}
where $x_0 \sim p_0$ and $x_1 \sim p_{\text{data}}$. The neural velocity field $v_\theta$ is trained by regressing toward the closed-form target velocity:
\begin{equation}
\mathcal{L}_{\mathrm{FM}}
=
\mathbb{E}_{x_0,x_1,t}
\left[
\| v_\theta(x_t,t) - (x_1 - x_0) \|_1
\right].
\end{equation}
Integrating the probability flow ODE transports a prior sample to the data manifold at $t=1$. This deterministic transport formulation removes the need for stochastic denoising and serves as the basis for conditional acoustic modeling.

\subsection{NeuroSonic}
\myparagraph{Conditional Acoustic Transport.}
We cast EEG-to-speech reconstruction as conditional transport of acoustic trajectories. Given paired EEG–audio samples $(E,X)$, we construct a corrupted acoustic state:
\begin{equation}
z_t = tX + (1-t)\varepsilon,
\qquad
\varepsilon \sim \mathcal{N}(0,I),
\end{equation}
and learn a velocity field that transports $z_t$ toward clean speech under EEG conditioning.
An overview of the architecture is shown in Fig.~\ref{fig:overview}. EEG and corrupted audio tokens are jointly processed to predict the probability-flow ordinary differential equation governing acoustic evolution. At inference, the learned ODE is integrated from $t=0$ to $t=1$ using a fixed-step Heun solver, yielding deterministic reconstruction conditioned on neural activity.

\myparagraph{Multimodal Tokenization and Alignment.}
Let $E \in \mathbb{R}^{C \times T_1}$ and $X \in \mathbb{R}^{T_2}$.  
EEG and acoustic signals are partitioned into non-overlapping patches:
$E \in \mathbb{R}^{C \times T_1}$ and $X \in \mathbb{R}^{T_2}$.
Each patch is projected into a shared latent space:
\begin{equation}
e_i = f_E(\mathrm{vec}(E_i)), 
\qquad
x_j = f_A(\mathrm{vec}(X_j)),
\end{equation}
with embedding dimension $d$.
Learnable modality embeddings and positional encodings are incorporated:
\begin{equation}
\tilde e_i = e_i + \tau_E + p_i,
\qquad
\tilde x_j = x_j + \tau_A + p_j.
\end{equation}
The resulting sequence:
\begin{equation}
Z = [\{\tilde e_i\}; \{\tilde x_j\}]
\in \mathbb{R}^{(N_E+N_A)\times d}
\end{equation}
enables global cross-modal interaction. Aggregating information through self-attention implicitly attenuates localized motion artifacts and low-SNR perturbations in scalp EEG.

\myparagraph{Time-Conditioned Gated Transformer.}
The sequence $Z$ is processed by $L$ pre-normalized Transformer blocks conditioned on interpolation time $t$:
\begin{align}
Z' &= Z + g_{\text{msa}} \cdot 
\mathrm{MSA}(\mathrm{AdaLN}(Z; t)), \\
Z  &= Z' + g_{\text{mlp}} \cdot 
\mathrm{MLP}(\mathrm{AdaLN}(Z'; t)).
\end{align}
where $\mathrm{AdaLN}(U;t)=\gamma_t\odot\mathrm{LN}(U)+\beta_t$, with $\gamma_t,\beta_t$ from the time embedding.
Global multi-head self-attention is defined as
\begin{equation}
\mathrm{MSA}(Z)
=
\mathrm{Concat}_{i=1}^{h}
\left(
\mathrm{Softmax}
\left(
\frac{Q_i K_i^\top}{\sqrt{d_h}}
\right)
V_i
\right) W^O,
\end{equation}
where $d_h = d/h$.
Time-dependent interpolation induces feature distribution shifts across $t$, potentially destabilizing attention logits~\cite{yang2026stable}. To control this effect, we apply per-head RMS normalization to query and key:
\begin{equation}
Q \leftarrow \mathrm{RMSNorm}(Q),
\qquad
K \leftarrow \mathrm{RMSNorm}(K).
\end{equation}
The network outputs $X_{\text{pred}} = \mathrm{net}(z_t,t,E)$, from which velocities are derived.

\myparagraph{Velocity-Based Objective.}
Under the manifold assumption~\cite{chapelle2006semi}, clean acoustic signals lie on a low-dimensional structure. Rather than regressing waveforms directly, we supervise transport dynamics in velocity space~\cite{li2026basicsletdenoisinggenerative}. Given
$\varepsilon = \frac{z_t - tX}{1-t}$, and $v_t = \frac{X - z_t}{1-t}$,
the predicted velocity is
$v_{\text{pred}} = \frac{X_{\text{pred}} - z_t}{1-t}$.
The final objective is
\begin{equation}
\mathcal{L}
=
\mathbb{E}_{X,\varepsilon,t}
\left[
\| v_{\text{pred}} - v_t \|_1
\right].
\end{equation}
Supervising transport in velocity space anchors learning on clean acoustic states and improves robustness under low-SNR and heterogeneous neural conditioning.

\section{Experiments}
\begin{table}[t]
\centering
\caption{Objective evaluation of EEG-conditioned speech reconstruction under cross-subject evaluation. Lower values indicate better performance for FAD, LSD, SC, and inference time (seconds). Results are reported as mean $\pm$ standard deviation. Best values for each metric are shown in bold.}
\vspace{-5pt}
\label{resultmel}
\renewcommand\arraystretch{0.95}
\setlength{\tabcolsep}{2mm}
\begin{tabular}{c|c|cccc}
\hline\hline
\multirow{2}{*}{Dataset} 
& \multirow{2}{*}{Method} 
& \multicolumn{4}{c}{Metrics} \\
\cline{3-6}
& 
& FAD~($\downarrow$) 
& LSD~($\downarrow$) 
& SC~($\downarrow$)
& Time (s)~($\downarrow$)
\\
\hline
\multirow{4}{*}{Cine}
& MF
  & 173.65  $\pm$ 0.26
  & 69.78 $\pm$ 0.11
  & 1.34 $\pm$ 0.09
  & 0.04
\\
& GAN 
  & 57.12 $\pm$ 3.49
  & 15.13 $\pm$ 0.14
  & 1.25 $\pm$ 0.11
  & \textbf{0.02}
\\
& DM  
  & 72.56 $\pm$ 1.69
  & 22.08 $\pm$ 0.07
  & 1.12 $\pm$ 0.04
  & 2.00
\\
& \textbf{Ours} 
& \textbf{39.06 $\pm$ 0.52} 
& \textbf{14.24 $\pm$ 0.31} 
& \textbf{0.64 $\pm$0.04 } 
& 0.86
\\
\hline
\multirow{4}{*}{EAV}
& MF  
& 85.27 $\pm$ 0.80
& 29.25 $\pm$ 0.64
& 1.49 $\pm$ 0.06
& 0.04\\
& GAN 
& 39.47 $\pm$ 0.83  
& 15.71 $\pm$ 0.34  
& 1.00 $\pm$ 0.01 
& \textbf{0.02}
\\
& DM  
& 15.87 $\pm$ 6.78 
& 19.47 $\pm$ 0.25 
& 1.25 $\pm$ 0.10 
 & 2.08
\\
& \textbf{Ours} 
& \textbf{11.64 $\pm$ 1.17} 
& \textbf{12.98 $\pm$ 0.16} 
& \textbf{0.28 $\pm$ 0.02} 
& 1.40
\\
\hline\hline
\end{tabular}
\end{table}

\subsection{Dataset}
We evaluate NeuroSonic on two publicly available EEG--audio datasets that span controlled conversational recordings and naturalistic audiovisual stimulation. After preprocessing, the combined corpus contains data from 48 subjects, totaling approximately 60 hours of synchronized EEG-audio recordings (49{,}200 paired segments).
CineBrain~\cite{gao2025cinebrainlargescalemultimodalbrain} provides simultaneously recorded EEG and fMRI during continuous audiovisual presentation. The accompanying audio includes speech as well as environmental sounds, yielding acoustically complex reconstruction targets. We follow the original protocol to temporally align EEG signals with the audio stream and reorganize continuous recordings into matched segments.
EAV~\cite{lee2024eav} consists of conversational interactions with synchronized EEG, audio, and video from 42 participants. Compared with CineBrain, EAV contains cleaner speech structure but stronger subject-specific variability arising from spontaneous dialogue and articulation differences.

For both datasets, preprocessing strictly follows the setting in~\cite{gao2025cinebrainlargescalemultimodalbrain,lee2024eav}. EEG signals undergo standard artifact removal procedures, including MRI-related artifact correction when applicable, 0.1--30\,Hz band-pass filtering, 50\,Hz notch filtering, and ICA-based removal of ocular, muscular, and cardiac components. All reported results are obtained under cross-subject evaluation, ensuring that test subjects are not observed during training.

\begin{figure}[t]
\centering
\includegraphics[width=0.98\linewidth]{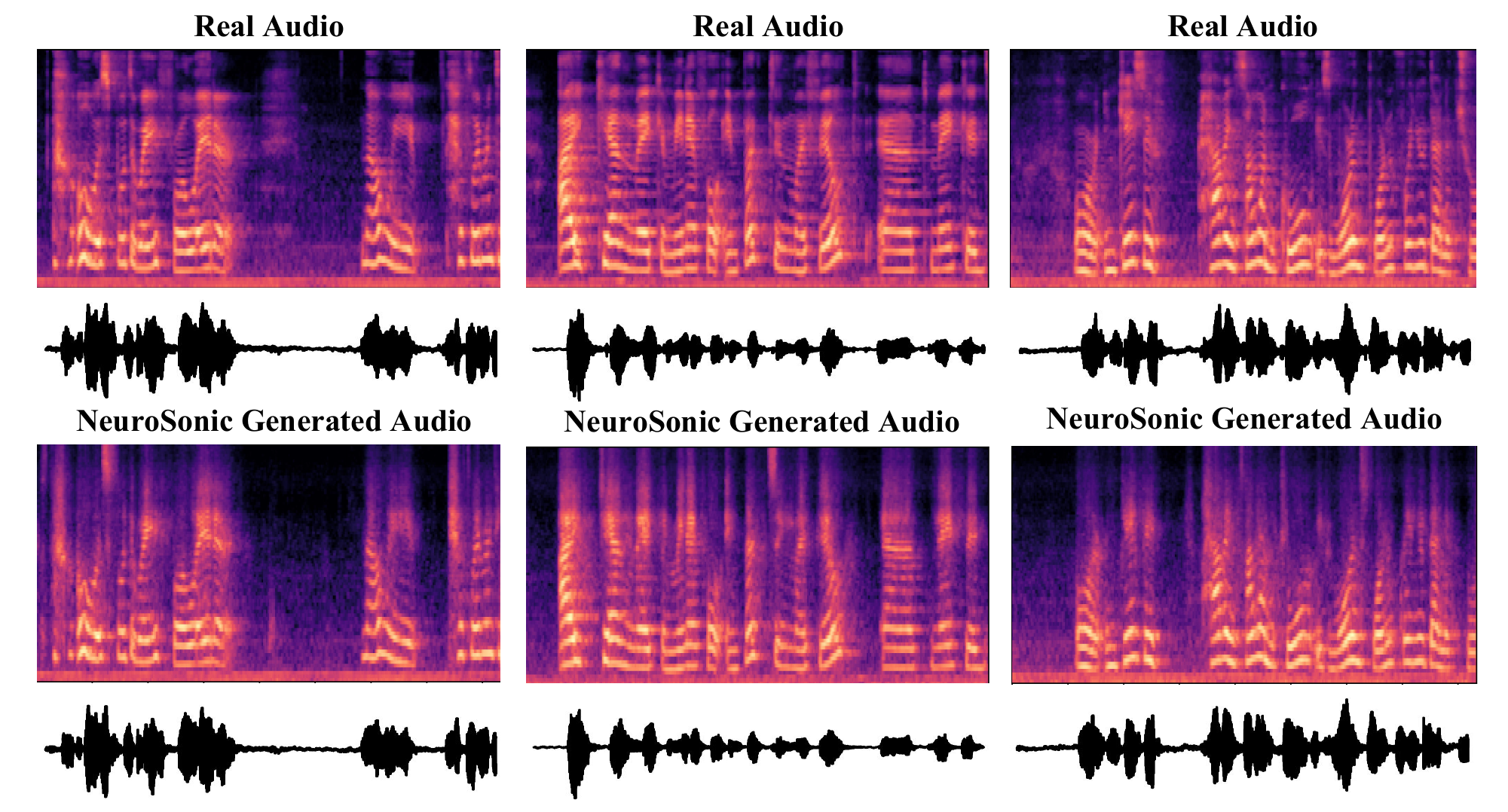}
\vspace{-5pt}
\caption{Comparison of ground-truth speech and NeuroSonic reconstructions. For each example, the reference mel-spectrogram and waveform are shown on top, with the EEG-conditioned reconstruction below. The reconstructed signals exhibit coherent formant trajectories and temporal modulation patterns consistent with the reference.}
\vspace{-5pt}
\label{fig:waveform}
\end{figure}

\subsection{Implementation Details}
\myparagraph{Setup.}
NeuroSonic uses a multimodal Transformer with 16 blocks, hidden size 1024, and 16 attention heads. Each block employs RMS-normalized self-attention and a gated MLP with a $4\times$ expansion ratio. To improve robustness to motion artifacts without over-regularizing early feature formation, dropout is applied selectively: attention, projection, and feed-forward dropout are enabled only in the middle blocks, while the earliest and latest blocks remain dropout-free.
Models are trained for 400 epochs with batch size 32 using AdamW, cosine learning-rate scheduling, and EMA tracking on an NVIDIA GeForce RTX5090 (32\,GB). Inference integrates the learned probability-flow ODE using 100 fixed Heun steps. Dataset-specific window lengths and channel counts follow the original preprocessing protocols~\cite{gao2025cinebrainlargescalemultimodalbrain,lee2024eav}.

\begin{table}[t]
\centering
\caption{Perceptual evaluation using DNSMOS. Higher values indicate better perceptual quality. Results are reported as mean $\pm$ standard deviation. Best results among learned models are shown in bold.}
\vspace{-5pt}
\label{resultmos}
\renewcommand\arraystretch{0.95}
\setlength{\tabcolsep}{2mm}
\begin{tabular}{c|c|ccc}
\hline\hline
\multirow{2}{*}{Dataset} 
& \multirow{2}{*}{Method} 
& \multicolumn{3}{c}{DNSMOS} \\
\cline{3-5}
& 
& SIG~($\uparrow$)
& BAK~($\uparrow$)
& OVRL~($\uparrow$)
\\
\hline
\multirow{5}{*}{Cine}
& GT  & 2.41 & 1.75 & 1.67 \\
& MF  
& 1.20  $\pm$ 0.01
& 1.10  $\pm$ 0.01
& 1.10  $\pm$ 0.01
  \\
& GAN 
  & 1.26 $\pm$ 0.03
  & 1.29 $\pm$ 0.01
  & 1.14 $\pm$ 0.01
  \\
& DM  
  & 1.21 $\pm$ 0.00
  & 1.33 $\pm$ 0.01
  & 1.07 $\pm$ 0.00
\\
& \textbf{Ours} 
  & \textbf{1.95 $\pm$ 0.04} 
  & \textbf{1.64 $\pm$ 0.04}
  & \textbf{1.44 $\pm$ 0.01}
  \\
\hline
\multirow{5}{*}{EAV}
& GT  & 3.32 & 2.77 & 2.47 \\
& MF  
& 1.19  $\pm$ 0.01
& 1.12  $\pm$ 0.01
& 1.08  $\pm$ 0.01
  \\
& GAN 
  & 1.98 $\pm$ 0.01 
  & 2.62 $\pm$ 0.05 
  & 1.45 $\pm$ 0.01 
  \\
& DM  
  & 2.92 $\pm$ 0.04 
  & 2.95 $\pm$ 0.10
  & 2.29 $\pm$ 0.11
\\
& \textbf{Ours} 
  & \textbf{3.31 $\pm$ 0.02} 
  & \textbf{3.07 $\pm$ 0.04}
  & \textbf{2.59 $\pm$ 0.03}
  \\
\hline\hline
\end{tabular}
\end{table}

\myparagraph{Baselines and Evaluation.}
We compare NeuroSonic with three representative generative paradigms for continuous signal synthesis: GANs~\cite{kong2020hifi}, diffusion models~\cite{schneider2023mousaitexttomusicgenerationlongcontext}, and mean flows~\cite{geng2025meanflowsonestepgenerative}. Each baseline is adapted to EEG conditioning in the simplest direct form: the GAN generator maps EEG features to waveform outputs; the diffusion model introduces EEG embeddings through cross-attention; and the mean-flow baseline concatenates EEG temporal embeddings as global conditioning.
Reconstruction quality is evaluated from four complementary perspectives: distributional alignment via Fréchet Audio Distance (FAD$\downarrow$), spectral fidelity via Log-Spectral Distance (LSD$\downarrow$) and Spectral Convergence (SC$\downarrow$), inference time (seconds$\downarrow$), and perceptual quality using DNSMOS$\uparrow$~\cite{reddy2021dnsmos}. Together, these metrics reflect statistical realism, harmonic structure, computational efficiency, and subjective intelligibility.

\begin{figure}[t]
\centering
\includegraphics[width=0.98\linewidth]{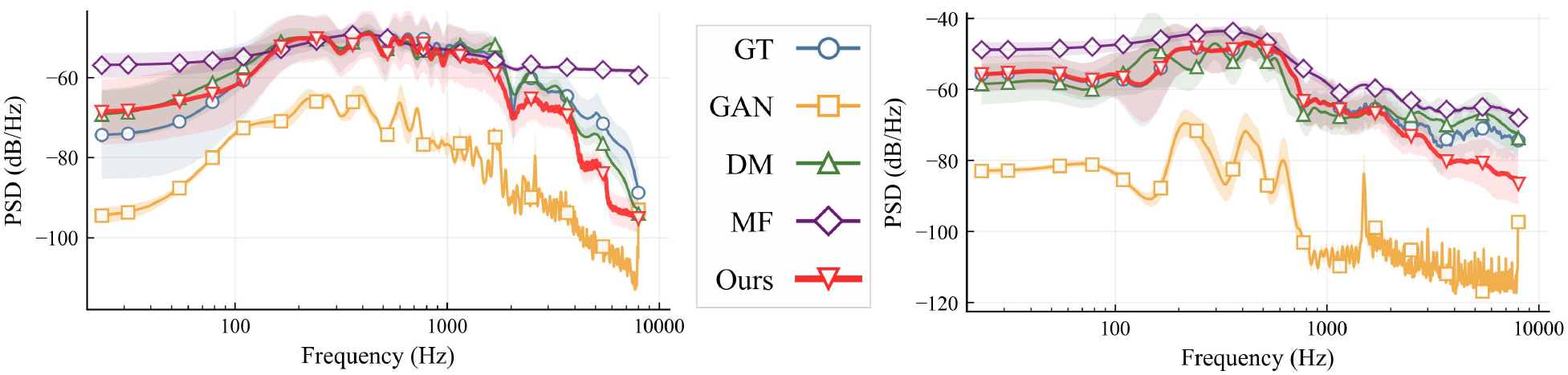}
\vspace{-5pt}
\caption{Power spectral density (PSD) of reconstructed audio on the Cine dataset (\textit{left}) and the EAV dataset (\textit{right}). Ground-truth audio (GT) is shown in blue. NeuroSonic (red) more closely follows the ground-truth spectrum in the low-frequency band and maintains consistent spectral behavior across datasets. GAN outputs exhibit broader spectral deviations, while diffusion models show increased energy in higher-frequency regions.}
\label{fig:PSD}
\vspace{-5pt}
\end{figure}

\subsection{Result}
\myparagraph{Distributional and spectral fidelity.}
Table~\ref{resultmel} reports objective reconstruction quality. NeuroSonic attains the best FAD and LSD on both datasets and yields a substantial reduction in spectral convergence error, indicating improvements that go beyond matching marginal audio statistics and extend to fine-grained spectral structure. The advantage is most pronounced on Cine, whose audio contains substantial background content and heterogeneous acoustic events, where stochastic baselines are more affected by artifact- and subject-dependent conditioning.

Fig.~\ref{fig:PSD} further illustrates this trend: NeuroSonic aligns most closely with the ground-truth PSD in the low-frequency band that dominates perceived speech quality, while avoiding the high-frequency over-emphasis observed in diffusion baselines and the broadband distortion typical of GAN outputs.

\begin{table}[t]
\centering
\caption{Ablation study of clean-state velocity supervision. The $x$-loss variant replaces velocity supervision with direct waveform regression. Lower values indicate better performance for FAD, LSD, and SC; higher values indicate better perceptual quality. Best values per metric are shown in bold.}
\vspace{-5pt}
\label{ablation}
\renewcommand\arraystretch{0.95}
\setlength{\tabcolsep}{2mm}
\begin{tabular}{c|c|cccccc}
\hline\hline
\multirow{2}{*}{Dataset} 
& \multirow{2}{*}{Method} 
& \multicolumn{6}{c}{Metrics} \\
\cline{3-8}
& 
& FAD~($\downarrow$)
& LSD~($\downarrow$)
& SC~($\downarrow$)
& SIG~($\uparrow$)
& BAK~($\uparrow$)
& OVRL~($\uparrow$)
\\
\hline

\multirow{2}{*}{Cine}
& $x$-loss 
  & \textbf{32.23} 
  & 14.27 
  & 0.91
  & 1.45 
  & 1.30 
  & 1.18 
  \\
& \textbf{Ours} 
  & 39.06 
  & \textbf{14.24} 
  & \textbf{0.64}
  & \textbf{1.95} 
  & \textbf{1.64}
  & \textbf{1.44}
  \\
\hline

\multirow{2}{*}{EAV}
& $x$-loss 
  & {12.14}
  & 13.45 
  & 0.90 
  & 3.07
  & 2.66 
  & 2.28 
  \\
& \textbf{Ours} 
  & \textbf{11.64}
  & \textbf{12.98} 
  & \textbf{0.28}
  & \textbf{3.31} 
  & \textbf{3.07}
  & \textbf{2.59}
  \\
\hline\hline
\end{tabular}
\end{table}

\myparagraph{Human-perceptual quality.}
Table~\ref{resultmos} summarizes DNSMOS scores~\cite{reddy2021dnsmos}. NeuroSonic achieves the highest SIG and OVRL on both datasets and shows consistent improvement in BAK, suggesting that the reconstructions reduce background interference while preserving intelligible speech structure. On EAV, the reconstruction slightly exceeds the recorded reference in OVRL, driven primarily by higher BAK (3.07 vs.\ 2.77). A plausible explanation is that EEG reflects neural representations related to speech intent and perception rather than the full acoustic mixture: conditioning on EEG provides weak support for non-linguistic background components, effectively suppressing them in the generated waveform.

Qualitative examples in Fig.~\ref{fig:waveform} are consistent with the perceptual gains: NeuroSonic preserves coherent formant trajectories and temporal modulation patterns without the over-smoothing artifacts typically introduced by regression-style objectives.


\myparagraph{Ablation.}
Table~\ref{ablation} studies the effect of clean-state velocity formulation. The direct $x$-loss variant achieves competitive FAD, suggesting that endpoint regression can approximate coarse distributional properties. However, it consistently degrades LSD, SC, and all DNSMOS components across both datasets, indicating weaker harmonic organization and temporal coherence. This divergence highlights a key distinction: matching endpoints does not constrain the path from noisy to clean states, whereas velocity supervision explicitly trains the transport dynamics. By anchoring learning on clean-state velocities, NeuroSonic enforces a coherent evolution along the acoustic manifold, which is reflected in improved spectral structure and higher perceived quality.

\section{Conclusion}
We presented NeuroSonic, a conditional flow-matching approach to reconstruct continuous speech from scalp EEG. By reframing EEG-to-speech reconstruction as deterministic conditional transport, the model learns a probability-flow velocity field that maps corrupted acoustic states to clean speech in a single ODE integration, avoiding stochastic sampling chains that are prone to artifact- and subject-dependent variability. Across two public datasets under cross-subject evaluation, NeuroSonic improves distributional realism, spectral fidelity, and perceptual quality over representative GAN-, diffusion-, and mean-flow baselines. The ablation study further shows that clean-state velocity supervision is essential for preserving spectro-temporal structure, even when endpoint regression can match coarse statistics. These results suggest that trajectory-based conditional transport is a principled and stable direction for neural speech reconstruction, and motivates future work on richer linguistic objectives and broader naturalistic settings.

\noindent\textbf{Disclosure of Interests.} The authors declare that they have no competing interests related to this work.

%
%
%
\bibliographystyle{splncs04}
\bibliography{miccai.bib}
%




\end{document}